\documentclass{egpubl}
\usepackage{eg2024}
\ConferencePaper        
\CGFccby   

\usepackage[T1]{fontenc}
\usepackage{dfadobe}  

\usepackage{cite}  
\BibtexOrBiblatex

\electronicVersion
\PrintedOrElectronic

\ifpdf \usepackage[pdftex]{graphicx} \pdfcompresslevel=9
\else \usepackage[dvips]{graphicx} \fi

\usepackage{egweblnk}

\usepackage{enumitem}
\usepackage{amsmath}
\usepackage{amsfonts}
\usepackage{multirow}
\usepackage{bigdelim}
\usepackage{color}
\usepackage{booktabs}
\usepackage{ifthen}
\usepackage{hyperref}
\usepackage{array, makecell}
\usepackage{subcaption}
\usepackage{adjustbox}

\usepackage{bbold}
\usepackage[noabbrev,capitalise,nameinlink]{cleveref}
\creflabelformat{equation}{#2\textup{#1}#3}

\usepackage{microtype}

\usepackage{xcolor, pgf}
\usepackage{color, colortbl}
\usepackage[normalem]{ulem} 
\usepackage{placeins}
\usepackage{soul}
\usepackage[skins]{tcolorbox}

\usepackage{siunitx}

\newlength\myboxwidth
\usepackage{xr}
\makeatletter
\newcommand*{\addFileDependency}[1]{
  \typeout{(#1)}
  \@addtofilelist{#1}
  \IfFileExists{#1}{}{\typeout{No file #1.}}
}
\makeatother
\newcommand*{\myexternaldocument}[1]{%
    \externaldocument{#1}%
    \addFileDependency{#1.tex}%
    \addFileDependency{#1.aux}%
}
\myexternaldocument{supp_main_eg}

\definecolor{gray}{rgb}{0.5,0.5,0.5}
\definecolor{green}{rgb}{0, 0.6, 0}
\definecolor{orange}{rgb}{1, 0.5, 0}
\definecolor{mahogany}{rgb}{0.75, 0.25, 0.0}
\definecolor{purple}{rgb}{0.6, 0, 0.6}
\definecolor{darkgreen}{rgb}{0, 0.3, 0}
\definecolor{orange}{rgb}{1, 0.5, 0.}
\definecolor{lightblue}{rgb}{0.52, 0.75,0.91}
\definecolor{softgreen}{rgb}{0.66,0.87,0.74}
\definecolor{softred}{rgb}{0.86,0.43,0.56}
\definecolor{softblue}{rgb}{0.43,0.56,0.88}

\newcommand{\mode}{final}
\ifthenelse{\equal{\mode}{draft}}{
\colorlet{lightyellow}{yellow!50}
\newcommand{\minorhl}[1]{\sethlcolor{lightyellow}\hl{#1}}
\DeclareRobustCommand{\iccmt}[1]{
  \begingroup
  \definecolor{hlcolor}{RGB}{168,221,188}\sethlcolor{hlcolor}%
  \hl{\textbf{ichao:} #1}%
  \endgroup
}
\DeclareRobustCommand{\todo}[1]{
  \begingroup
  \definecolor{hlcolor}{RGB}{245,183,177}\sethlcolor{hlcolor}%
  \hl{\textbf{TODO:} #1}%
  \endgroup
}

\newcommand{\asc}[1]{\textcolor{green}{Arik:{#1}}}
\newcommand{\koyama}[1]{\textcolor{magenta}{koyama:\ {#1}}}
} {
\newcommand{\todo}[1]{}
\newcommand{\iccmt}[1]{}

\newcommand{\asc}[1]{}
\newcommand{\koyama}[1]{}
\newcommand{\minorhl}[1]{#1}
}

\newcommand{\ignore}[1]{}
\newcommand{\none}[1]{}
\newcommand{\com}[1]{}

\newcommand{\etal}{{\textit{et~al.}}}
\newcommand{\ie}{i.e.,}
\newcommand{\eg}{e.g.,}

\begin{document}

\title[FontCLIP: A Semantic Typography Visual-Language Model for Multilingual Font Applications]%
      {FontCLIP: A Semantic Typography Visual-Language Model for Multilingual Font Applications}

\author[Yuki Tatsukawa et al.]
{\parbox{\textwidth}{\centering Yuki Tatsukawa$^{1}$\orcid{0009-0003-5128-8032} ~ I-Chao Shen$^{1}$\orcid{0000-0003-4201-3793} ~ Anran Qi$^{1}$\orcid{0000-0001-7532-3451} ~ Yuki Koyama$^{2}$\orcid{0000-0002-3978-1444}  ~ Takeo Igarashi$^{3}$\orcid{0000-0002-5495-6441} ~ Ariel Shamir$^{4}$\orcid{0000-0001-7082-7845}
        }
        \\
{\parbox{\textwidth}{\centering $^1$ \{tatsukawa-yuki537, ichaoshen, annranqi1024\}@g.ecc.u-tokyo.ac.jp , The University of Tokyo, Japan \\
$^2$ koyama.y@aist.go.jp, National Institute of Advanced Industrial Science and Technology (AIST), Japan \\
$^3$ takeo@acm.org, The University of Tokyo, Japan \\
$^4$ arik@runi.ac.il, Reichman University, Israel
      }
}
}

\maketitle
\begin{abstract}
Acquiring the desired font for various design tasks can be challenging and requires professional typographic knowledge.
While previous font retrieval or generation works have alleviated some of these difficulties, they often lack support for multiple languages and semantic attributes beyond the training data domains. To solve this problem, we present FontCLIP – a model that connects the semantic understanding of a large vision-language model with typographical knowledge.
We integrate typography-specific knowledge into the comprehensive vision-language knowledge of a pretrained CLIP model through a novel finetuning approach.
We propose to use a compound descriptive prompt that encapsulates adaptively sampled attributes from a font attribute dataset focusing on Roman alphabet characters.
FontCLIP's semantic typographic latent space demonstrates two unprecedented generalization abilities.
First, FontCLIP generalizes to different languages including Chinese, Japanese, and Korean (CJK), capturing the typographical features of fonts across different languages, even though it was only finetuned using fonts of Roman characters.
Second, FontCLIP can recognize the semantic attributes that are not presented in the training data. 
FontCLIP's dual-modality and generalization abilities enable multilingual and cross-lingual font retrieval and letter shape optimization, reducing the burden of obtaining desired fonts.
\end{abstract}

\section{Introduction}
\label{sec:intro}

Acquiring a suitable font is a crucial step in many design workflow, especially when designing a poster or a banner with cross-lingual characteristics.
While previous works have facilitated font retrieval~\mbox{\cite{o2014exploratory,chen2019large}} and generation~\mbox{\cite{WangSIGGRAPH2020,wang2021deepvecfont}}, they are often limited to the languages and attributes presented in the training data.
The available datasets~\mbox{\cite{o2014exploratory,chen2019large}} only include Roman fonts and their associated attributes, which does not allow users to obtain fonts in other languages.
Furthermore, the current datasets have limited annotated attributes, which prevents users from specifying their desired fonts freely. 
\begin{figure}[h!]
\centering\includegraphics[width=\linewidth]{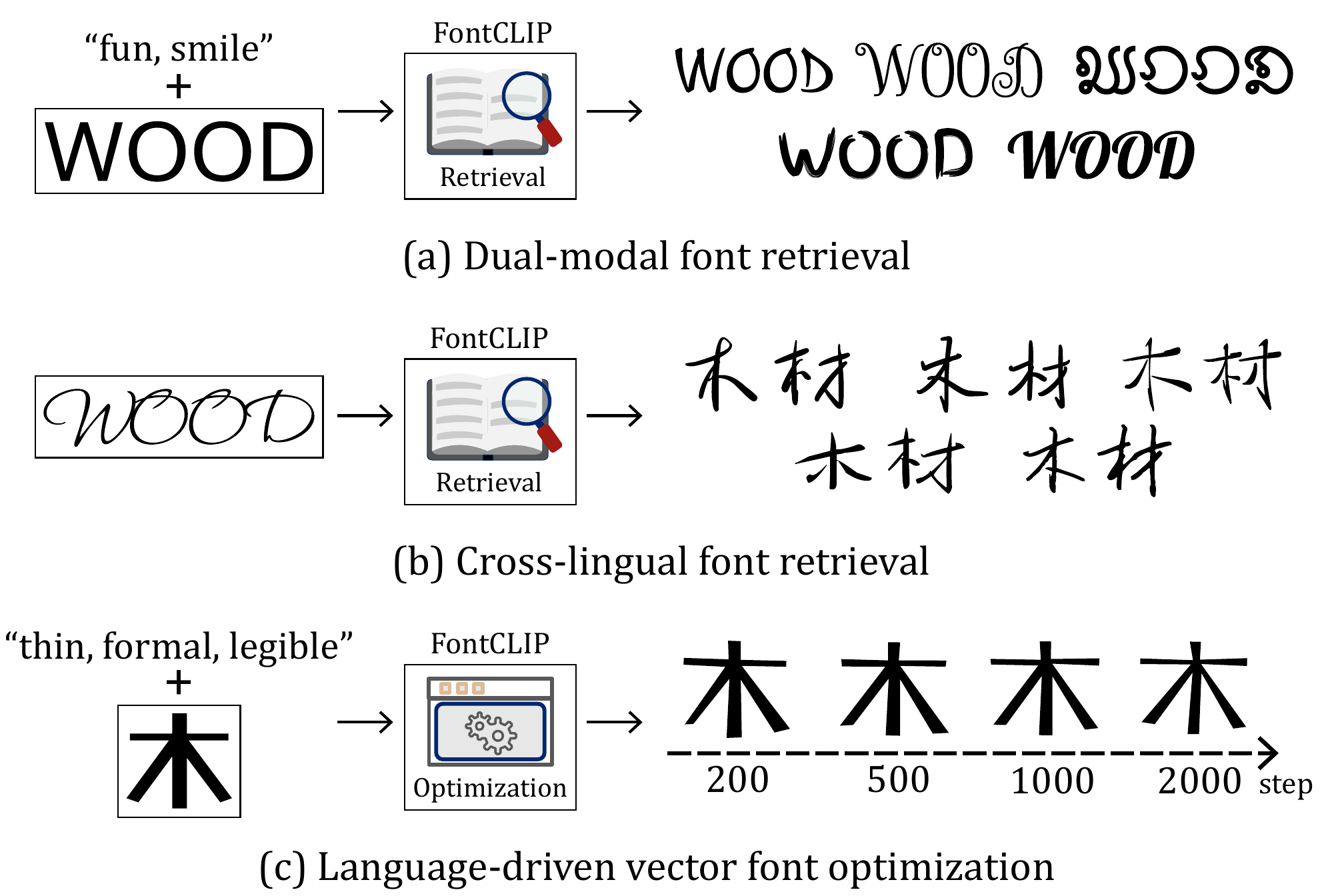}
  \caption{
    FontCLIP enables the following typography-specific applications.
    (a) \textbf{Dual-modal font retrieval}: our method retrieves results that preserve the style of the query font image (text with frame) while incorporating the desired attributes.
    (b) \textbf{Cross-lingual font retrieval}: our method retrieves results in other languages with a similar query font image style.
    (c) \textbf{Language-driven vector font optimization}: our method manipulates the shape of input letters aligning with a set of desired attributes.    
  }
  \label{fig:teaser}
\end{figure}

In this paper, we tackle these deficiencies by defining a semantic latent space connecting language and visuals to \emph{typography} to enable multilingual and cross-lingual font retrieval and editing tasks.
We base our technique on modifying a pretrained vision-language model trained on large-scale natural image and text pair, and without requiring any additional data-gathering beyond what is already available. 
Models such as CLIP (Contrastive Language–Image Pre-training)~\cite{radford2021learning}, have demonstrated exceptional capabilities in learning aligned visual and language features.
CLIP has exhibited remarkable zero-shot recognition capabilities, empowering a wide range of downstream visual recognition tasks ~\cite{kuo2023open,zhou2022zegclip,luo2022segclip,zhou2022extract}.
Additionally, the latent space of CLIP has been used for various content generation applications, including images~\cite{ramesh2022hierarchical}, abstract sketches~\cite{vinker2022clipasso}, 3D avatars~\cite{hong2022avatarclip}, and artistic images~\cite{rombach2021highresolution}. 
These works collectively highlight that CLIP's latent space carries profound semantic understanding that can connect between language and visuals.

\begin{figure}[!t]
  \centering
  \includegraphics[width=\columnwidth]{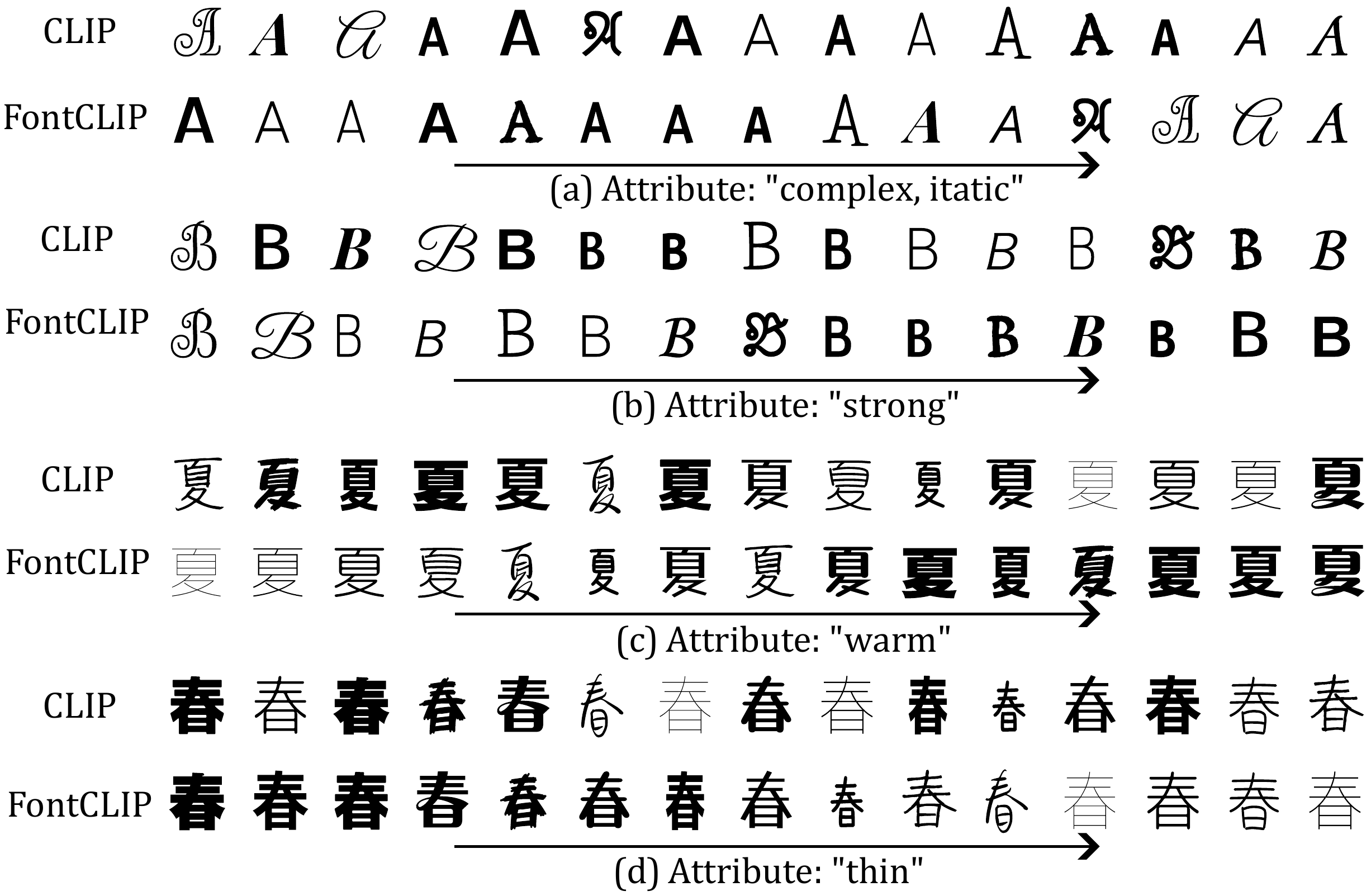}
  \caption{
    Glyph images of Roman and Chinese letters sorted by (a) ``complex, italic'', (b) ``strong'', (c) ``warm'', and (d) ``thin'' attribute scores predicted by CLIP and FontCLIP. FontCLIP's sorting aligns more closely with human perception.
  }
  \label{fig:motivation}
\end{figure}

However, typography is a very specialized domain that is different from natural photographs, paintings, or sketches, which were originally used to train CLIP. 
Hence, simply using the original CLIP model cannot effectively recognize visual typographic characteristics and establish meaningful connections with language representations, as illustrated in~\cref{fig:motivation}.
The primary reason for this is the substantial domain disparity between the typographical image and those portraying natural scenes.
Moreover, the language used to describe typographic data often diverges from the descriptions of natural scenes represented in different styles. 

We present \textit{FontCLIP}, a CLIP-based model specifically designed to learn a semantic typographic latent space that bridges language and visual typographic attributes, enabling various typographic applications.
By finetuning a pretrained CLIP model on font data with attribute scores, we enhance its zero-shot recognition capability and enable it to generalize to the typography domain.
The learned features of FontCLIP enable prediction of multilingual visual typographic attributes with the ability to generalize to \emph{out-of-domain} attributes.
Remarkably, we achieve these generalizations by using an existing Roman character dataset without the need for collecting any new data.

Finetuning FontCLIP is accomplished using a novel compound descriptive prompt that encapsulates multiple attributes within a single prompt.
To determine the attributes included in the compound descriptive prompt, we adaptively sample them to cover the distribution of attribute scores and convert continuous score into sampled text.
In our finetuning process, we use a randomly generated compound descriptive prompt and a font image with a random augmentation transformation at each iteration, thereby significantly expanding the original font attributes dataset~\cite{o2014exploratory}.

We evaluated the performance of FontCLIP through quantitative experiments that assessed the correlation between the predicted and manually annotated attribute scores. 
Our experiments reveal that the finetuned FontCLIP model, originally trained on Roman alphabet characters with $37$ attributes, exhibits unprecedented generalization capabilities.
First, FontCLIP is capable of generalizing to \emph{out-of-domain} languages, which makes it possible to use it for multilingual and cross-lingual font-related tasks.
Second, it can generalize to \emph{out-of-domain} attributes, which means it can be used for font retrieval and editing using language descriptions beyond the original attribute set. 
Lastly, by leveraging the dual-modality of CLIP, FontCLIP allows users to obtain the desired fonts through both desired text attributes and font image examples.

We demonstrate FontCLIP in two major applications: (1) a novel dual-modal font retrieval interface that surpasses the traditional drop-down list interface in terms of user satisfaction and achieves similar performance without using vector-based typographical features extracted from all Roman characters and, (2) a novel optimization framework that utilizes FontCLIP latent space to manipulate vector letter shapes based on desired attributes or font image examples, thereby opening up exciting possibilities for font customization (see~\cref{fig:teaser}).

To sum up, we make the following contributions:
\begin{itemize}
    \item To the best of our knowledge, we present the first visual-language model that learns a semantic typographic latent space. 
    Through experiments and user studies, we validate its generalization abilities over multilingual and \emph{out-of-domain} attributes.
    \item We present a novel approach to finetune a vision-language model using font data with attribute scores.
    \item We present a dual-modal font retrieval application based on FontCLIP that uses visual examples and language descriptions to search for appropriate fonts across different languages.
    \item We present an optimization-based method to modify the shape of letters in vector representation to better match either a set of language descriptions or a visual input image sample.
\end{itemize}

\section{Related Work}
\label{sec:related}

\subsection{Vision-Language Representations and Applications}
Traditional visual recognition models are often constrained in their practicality since they are trained to recognize a predetermined set of object categories.
Hence, they require additional labeled data to generalize to new visual concepts and domains.
However, recent advancements in large vision-language models pretrained on vast image-text pairs have demonstrated that such models can acquire rich image and object-level visual representations~\cite{radford2021learning,jia2021scaling,li2022grounded}.
These models are semantically rich because the paired texts contain a broader set of visual concepts than any pre-defined concept set.
Thus, the learned representations can be directly used for downstream image recognition tasks such as image classification~\cite{radford2021learning,jia2021scaling}, object recognition~\cite{kuo2023open}, image segmentation~\cite{zhou2022zegclip,luo2022segclip,zhou2022extract}, and text-image retrieval~\cite{liu2021image} in zero-shot setting.
Moreover, the learned representations are applied to 3D shape classification~\cite{zhang2022pointclip}, 3D part segmentation~\cite{liu2022partslip}, 3D avatar and shape generation~\cite{hong2022avatarclip,michel2022text2mesh,mohammad2022clip,Gao2023SIGGRAPH,tevet2022motionclip} and NeRF generation and manipulation~\cite{jain2022zero,wang2022clip} tasks.
To the best of our knowledge, we propose the first semantic typography visual-language model that connects language and visual typographic attributes.
By doing so, FontCLIP enables font retrieval and manipulation tasks with a broader range of semantic concepts, expanding the possibilities for font customization and exploration.

\subsection{Font Retrieval and Interface}
Font selection is the process of selecting fonts from a set of fonts based on user-specified conditions across different formats. 
When the desired fonts are presented as images, traditional visual font recognition approaches identify the typeface, weight, and slope of text within them~\cite{wang2015deepfont,chen2014large}.
In the context of graphic design, users frequently aim to find a font that complements the overall design elements. 
As a result, they often rely on traditional font selection interface, such as a long list of font names, which is overwhelming to navigate and utilize.
To address this issue, O'Donovan~\etal~\cite{o2014exploratory} proposed selecting fonts using semantic attributes and collected a font attribute dataset.
More recent advancements have introduced larger font attribute datasets and deep learning-based methods to improve the accuracy and efficiency of font retrieval~\cite{chen2019large, choi2019assist}.
However, these previous attribute-based font selection methods only work on in-the-domain attributes and scripts.
In contrast, using FontCLIP latent space enables font retrieval with out-of-domain attributes and scripts, thereby offering enhanced flexibility and efficiency in font selection.

\subsection{Vector Font Generation}
Example-based methods generate a complete character set of a font~\cite{suveeranont2010example} or a personalized handwritten style~\cite{chen2015data,lian2018easyfont} from a single character.
Parameterizing fonts is another method that allows users to create novel fonts by adjusting a set of parameters~\cite{Ariel1998Feature,knuth1982concept}.
Campell and Kautz~\shortcite{campbell2014learning} took a step further and presented the first generative model for fonts. 
By exploring the learned manifold, the model enables interpolation between existing fonts and the discovery of new fonts.
Recently, various deep learning-based methods have been proposed for synthesizing vector glyphs~\cite{lopes2019learned,carlier2020deepsvg,wang2021deepvecfont,reddy2021im2vec}.
However, these methods often require users to provide sample characters of the desired font, making them challenging if they lack such resources.
\minorhl{To address this issue, Wang\mbox{~\etal~\shortcite{WangSIGGRAPH2020}} proposed Attribute2Font, which generates glyph images solely based on user-specified attributes.
However, they can only generate bitmap glyph images for attributes and languages that are included in the training data.
Thus, to generate bitmap glyph images for new attributes or languages, more training data is necessary.
In contrast, our method can generate vector fonts that are easily manipulable.
Moreover, our method can generate vector fonts for attributes and languages that are not part of the training data without requiring additional training data.
}

\minorhl{
Our vector font optimization method is inspired by Word-As-Image\mbox{~\cite{iluz2023wordasimage}}, but with two significant differences.
First, while Word-As-Image focuses on deforming a vector letter toward a conceptual visual representation (\eg~cat or dog), our optimization method using FontCLIP concentrates on capturing and reconstructing typographical features of each character (\eg~thin, italic, and serif).
As we demonstrated in \mbox{~\cref{fig:opt_text_driven}}, our method is more effective at capturing and reconstructing typographical features than Word-As-Image.
Second, our optimization method utilizes FontCLIP's text and image encoder, which allows for dual-modal font optimization.
In contrast, Word-As-Image only operates on text input. 
This makes our optimization method more practical and useful for capturing fonts in real-world scenarios, as shown in\mbox{~\cref{fig:opt_billboard}}.
}

\section{CLIP Preliminaries}
CLIP~\cite{radford2021learning} is a vision-language model pretrained on a large number of image-text paired data and trains both an image encoder $E_I$ and a text encoder $E_T$ to a joint latent space.
During training, CLIP uses a contrastive loss to learn a joint embedding for the two modalities. 
Specifically, for a mini-batch of image-text pairs, CLIP maximizes for each matching image-text pair the cosine similarity of their embeddings while minimizing the cosine similarities with all other unmatched texts/images.
After training, the joint latent space of CLIP enables various downstream image processing and vision tasks in a zero-shot manner.
For example, in image classification, given an input image $I$, its image embedding ($\mathbf{x}=E_I(I)$) is found using the image encoder, and a set of text embeddings ($\{\mathbf{w}_i=E_T(T_i)\}_{i=1}^{K}$) are found using the text encoder. 
In particular, each $\mathbf{w}_i$ is derived from a prompt $T_i$, such as ``a photo of a \{class\}'' where the ``\{class\}'' token is filled with the $i$-th class name.
The prediction probability of class $y$ is then defined as:
\begin{align}
p(y|\mathbf{x})=\frac{\text{exp}(\text{sim}(\mathbf{x},\mathbf{w}_y)/\tau)}{\sum_{i=1}^{K}\text{exp}(\text{sim}(\mathbf{x},\mathbf{w}_i)/\tau)}
\end{align}
where $\text{sim}(\cdot,\cdot)$ denotes cosine similarity and $\tau$ is a learned temperature parameter.

\section{F\lowercase{ont}CLIP}
\label{sec:fontclip}

\begin{figure}[t!]
\centering
\includegraphics[width=1.0\linewidth]{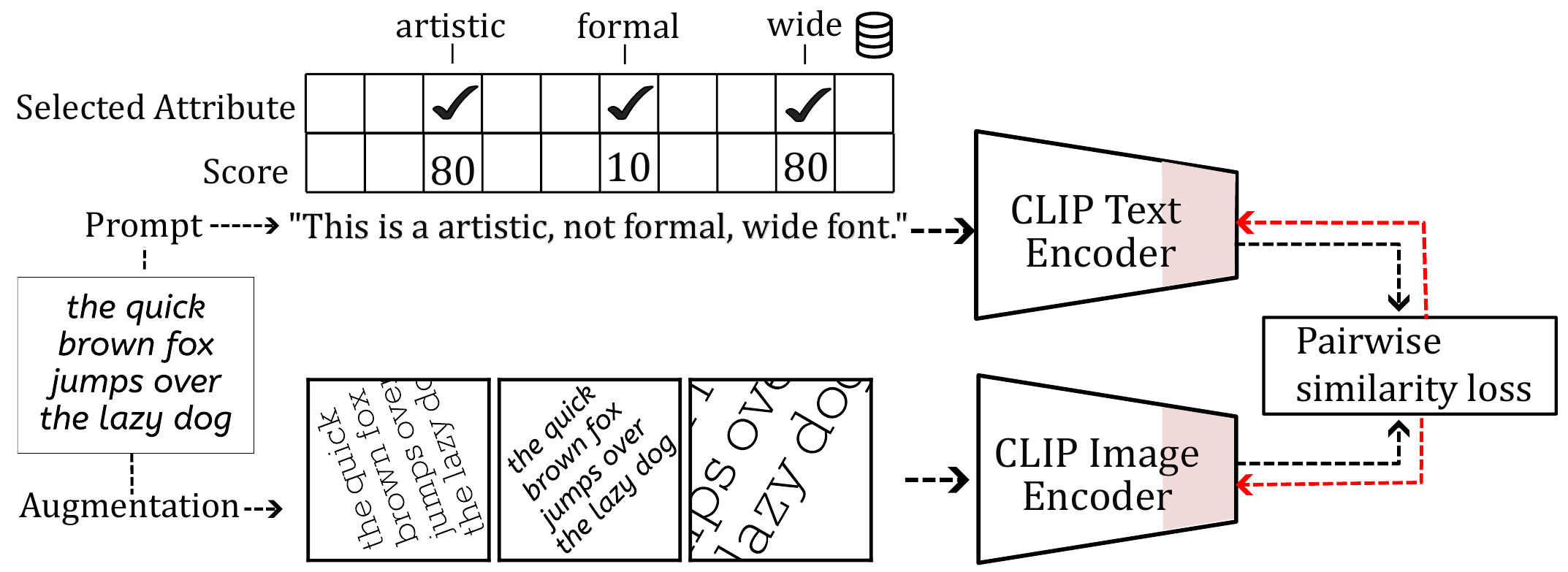}
  \caption{
    \textbf{Overview of FontCLIP finetuning.}
    During each finetuning iteration, we randomly select attributes based on their scores from an existing font-attribute dataset and create a compound descriptive prompt for each font.
    Simultaneously, we generate a font image and apply a random augmentation transformation to enhance variability.
    We finetuned the last three transformer blocks (highlighted in red) for both encoders using a pairwise similarity loss function.
  }
  \label{fig:finetune_overview}
\end{figure}
Our goal is to learn a semantic typography latent space that can be effectively used for various typographic applications, including font retrieval and optimization-based font manipulation.
To achieve this goal, we focus on incorporating typography-specific knowledge into the existing large pretrained vision-language model CLIP. 
Our approach is to finetune a pretrained CLIP model using pairs of descriptive prompts and font image as inputs derived from a font-attribute dataset (\cref{fig:finetune_overview}).
In the following, we provide the technical details of our finetuning approach as well as the rationale behind its design.

\subsection{Finetuning - Training Data}
For finetuning, we use the dataset from~\cite{o2014exploratory}.
This dataset consists of $200$ Roman fonts, each annotated with $37$ attribute scores.
The attributes in the dataset can be broadly classified into two categories.
Some of these attributes are related to the shape of the fonts, such as ``serif'', ``italic'', and ``thin''. 
The other attributes describe perceptual qualities such as ``friendly'', ``warm'', and ``happy''.
Each font in the dataset is assigned scores ranging from $0$ and $100$ for each attribute.
Among these attributes, there are some binary attributes: ``capitals'', ``cursive'', ``display'', ``italic'', ``monospace'' and ``serif'', meaning that they are assigned a score of either $0$ or $100$.
To suit our finetuning requirement, we modify the original dataset and create a prompt-based dataset that aligns better with our objectives.

\subsubsection{Compound Descriptive Prompt}
Radford~\etal~\shortcite{radford2021learning} introduced a hand-crafted prompt: ``a photo of a ``\{class\}'' for generic objects and scenes.
However, in the case of our font dataset, each font is characterized by multiple attributes simultaneously with continuous scores, which differs from the original classification task that the original CLIP model was trained on.
As a result, the above simple prompt is inadequate for accurately describing each font in our dataset.
To overcome this challenge, we propose to use a compound descriptive prompt, combined with an adaptive sampling technique, to generate a more comprehensive and descriptive prompt for each font in our dataset.

During each finetuning iteration $i$, we generate the compound prompt $T^F_i$ for a font $F$ as
\begin{align}
T^F_i = \text{``This is} ~[A]_1, [A]_2,..., [A]_N ~ \text{font.''},
\end{align}
where each $[A]_n$ represents the $n$-th sampled attribute and $N$ denotes the total number of attributes sampled.
Throughout the finetuning process, we randomly set $N$ to between $1$ and $3$ for each iteration.
To determine the expression for each attribute, we consider whether its score is over or below $50$.
Specifically, if the score is over $50$, we use the expression $[A]_n = \text{``[attribute]''}$.
Conversely, if the score is below $50$, we use $[A]_n = \text{``not [attribute]''}$.
For example, if the score of attribute ``happy'' for font $F$ exceeds $50$, the corresponding expression in the compound prompt is set to ``happy''.
Otherwise, it is set to ``not happy''.
We randomly selected attributes for each font based on their attribute score distribution.
Specifically, the probability of attribute $a$ being selected is computed as
\begin{align}
p(a) = \frac{\|S(a)-50\|}{\sum_{i=1}^{37}\|S(a_i)-50\|},
\end{align}
where $S(a)$ represents the score of attribute $a$.
We show how we generate a compound descriptive prompt in the top row of~\cref{fig:finetune_overview}.

\subsubsection{Font Image}
\label{sec:font_image}
To generate the training images for each font in the dataset, we adopt the same approach used by O'Donovan \etal~\shortcite{o2014exploratory}.
Specifically, we render an image of each font using the text ``\textit{The quick brown fox jumps over the lazy dog}'', which is widely used for displaying font samples due to its inclusion of a diverse range of letters~\cite{o2014exploratory,Tugba2020Happy}.
To enhance the robustness of our finetuned model, we apply standard data augmentation techniques such as rotation, cropping, and scaling to the font image (refer to \cref{fig:finetune_overview}).

\subsection{Finetuning - Loss Function}
Our finetuning method does not apply the CLIP-style contrastive learning~\cite{radford2021learning} directly.
Instead, it only requires positive pairs that include a compound descriptive prompt and a font image.
The reason for this is that negative expressions like $\text{``not [attribute]''}$ are already incorporated in our compound descriptive prompt.  
Thus, we can finetune the pretrained model to learn effective semantic typographic latent space solely by maximizing the cosine similarity between the embedded vectors of the compound descriptive prompts and those of their corresponding font images. Specifically, given a pair of a font image $I^F$ and a compound descriptive prompt $T^F$, we define the \textit{pairwise similarity} loss function as:
\begin{align}
    \mathcal{L}_{PS} = - \frac{1}{n}\sum_{q=1}^n{\frac{E_I(I^F_q) \cdot E_T(T^F_q)}{\lVert E_I(I^F_q)\rVert \lVert E_T(T^F_q)\rVert}},
    \label{eq:pos_loss}
\end{align}
where $n$ is the number of font descriptive prompt and font image pairs, $E_I$ and $E_T$ are the image and text encoder of the finetuned CLIP model, respectively.

\subsection{Implementation Details}
Our finetuning approach is based on the pretrained ViT CLIP model, which is publicly available on Hugging Face\footnote{\url{https://huggingface.co/sentence-transformers/clip-ViT-B-32}}.
Throughout the finetuning process, we update the weights of the last three transformer block layers in both text and image encoders for $3,000$ epochs, while keeping the remaining weights frozen.
We use the Adam optimizer \cite{kingma2015adam} with a learning rate of \num{2e-5}, which is halved every $500$ epochs.
The resolution of the font image used in finetuning is $ 214 \times 214$.
In our setting, the finetuning process took around $12$ hours on a machine equipped with an i7-12700K CPU with $32$GB memory and RTX3080 GPU with $10$GB memory.

\section{Experiments}
We conducted experiments using \textit{in-domain} and \textit{out-of-domain} attributes to evaluate the performance of FontCLIP.
Specifically, we measured the correlations between the attribute score predicted using the FontCLIP features and the ground truth attribute score using the dataset from~\cite{o2014exploratory}.

\subsection{In-Domain Attributes}
The first experiment aims to evaluate the consistency for \emph{in-domain} attributes, wherein all attributes are used during the finetuning of each model.
We used all $200$ fonts from \cite{o2014exploratory}, which we randomly divided into $140$ fonts for training, $30$ fonts for validation, and $30$ fonts for testing.
We finetuned FontCLIP using the $140$ fonts from the training set.
For each font $F$ in the testing dataset and for each attribute $[A]$, we calculated the similarity score between the font and the attribute in the following process.
First, we obtained the visual embedding vector $E_I(I_F)$ of the font visual prompt $I_F$ associated with $F$.
Next, from the attribute $[A]$, we created a descriptive prompt $T_A$: ``This is a $[A]$ font.'' and obtained the text embedding vector $E_{T}(T_A)$ for this prompt.
Finally, we calculated the cosine similarity between $E_I(I_F)$ and $E_{T}(T_A)$.
We considered a model's performance as the average correlation between predicted attribute scores and ground truth scores for all fonts in the testing set across all attributes.
We compared three models: FontCLIP without using compound descriptive prompts (CDP), FontCLIP with CDP, and the baseline CLIP model.

\begin{figure}[t!]
    \centering
    \includegraphics[width=\columnwidth]{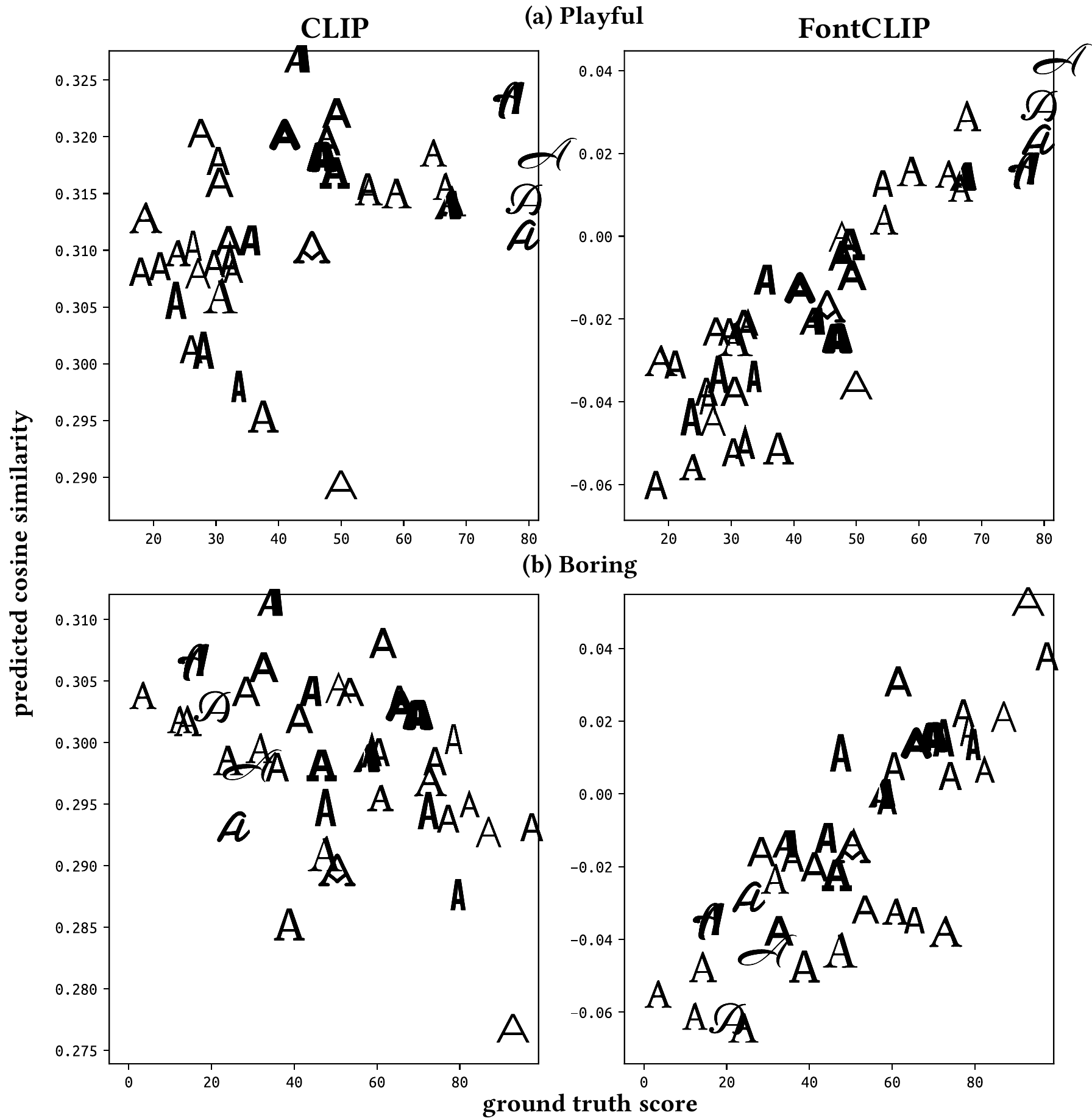}
    \caption{
    The visualization of the correlation between the predicted similarity scores from CLIP and FontCLIP, and the ground truth scores for (a) ``playful'' and (b) ``boring'' attributes.
    }
    \label{fig:correlation_visualization}
\end{figure}

As shown in the first row in~\cref{tab:correlation}, all variants of FontCLIP surpasses CLIP by a substantial margin.
This observation suggests that FontCLIP has effectively learned the relationship between visual typographic attributes and the corresponding semantic attributes described by language, resulting in attribute ratings that are more aligned with human ratings.
In addition, we provide a visualization of the correlation between predicted similarity scores and ground truth scores for the attributes ``thin'' and ``playful'' in \cref{fig:correlation_visualization}.
This visualization allows us to observe the alignment between the predicted attribute scores and the ground truth scores.
We also include correlation visualizations for other attributes in the supplemental material.

\begin{table}[t]
\centering
\begin{tabular}{ccc}
\toprule
Model & \textit{In-domain} $\uparrow$& \textit{Out-of-domain}$\uparrow$ \\
\midrule
CLIP & $0.159$ & $0.159$ \\ 
\makecell{FontCLIP (w/o CDP)} & $0.704$ & $0.317$  \\  
FontCLIP & $\mathbf{0.723}$ & $\mathbf{0.404}$  \\
\bottomrule
\end{tabular}
\caption{
The average correlations for \emph{in-domain} attributes and \emph{out-of-domain} attributes of CLIP, FontCLIP trained without using compound descriptive prompts (w/o CDP), and FontCLIP trained with CDP (ours).
By using CDP, FontCLIP can better generalize to \emph{out-of-domain} attributes.
}
\label{tab:correlation}
\end{table}

\subsection{Generalization to Out-of-Domain Attributes}
To evaluate the generalization capability of the FontCLIP latent space to \emph{out-of-domain} attributes, meaning they are absent in the finetune training data, we conducted a leave-one-out experiment.
During this experiment, we used all $200$ fonts as training data for finetuning the model but excluded one attribute at a time during the finetuning process.
Then, we calculated the average correlation between the predicted similarity score and ground truth attribute scores for all fonts, solely for the excluded attribute.
This process was repeated for each attribute in the dataset from \cite{o2014exploratory}, resulting in a total $N$ finetuning process, where $N$ represents the number of attributes. 
The performance of each model was evaluated by computing the average correlation across all $N$ attributes.

The results presented in the second row of~\cref{tab:correlation} highlight that FontCLIP (w/o CDP) already performs better than CLIP.
Moreover, the inclusion of CDP noticeably enhances the correlations.
These findings indicate that FontCLIP can effectively generalize to \emph{out-of-domain} attributes with straightforward finetuning and compound descriptive prompts.
Additionally, we provide the correlation scores of all attributes in the supplemental material.
\begin{figure}[t]
    \centering
    \includegraphics[width=\columnwidth]{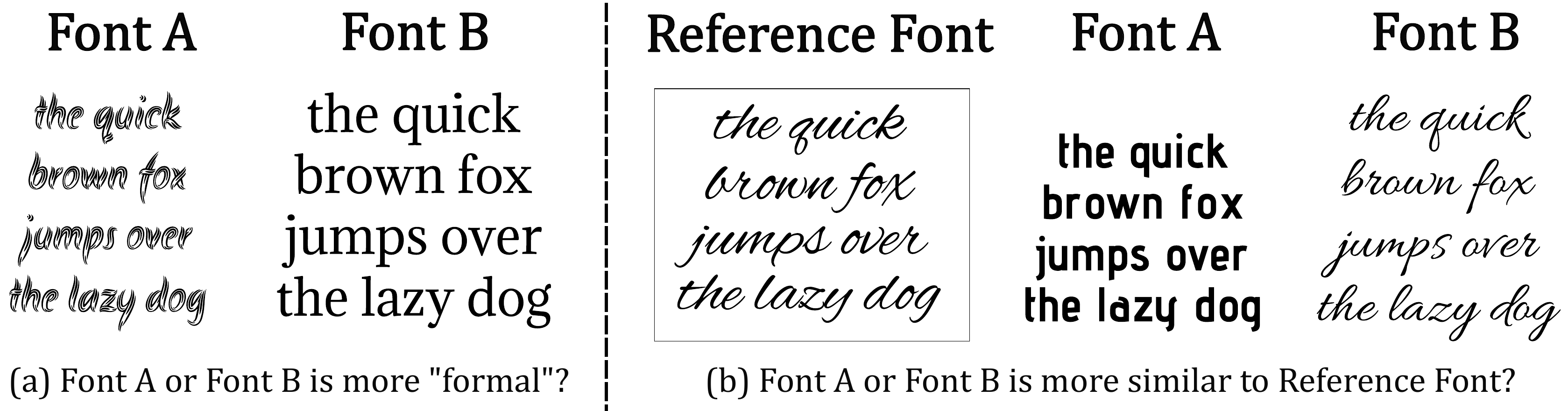}
    \caption{
    (a) In the pairwise attribute prediction task, we use a classifier to determine which font has a higher attribute score between two font options. 
    (b) In the pairwise similarity prediction task, we use a classifier to determine which font is more similar to a reference font between two font options.
    In both tasks, the obtained results are compared with human judgments, and the prediction accuracy is calculated as the performance metric.
    }
    \label{fig:classification_tasks}
\end{figure}

\section{Dual-Modal Multilingual Font Retrieval}
O'Donovan~\etal~\shortcite{o2014exploratory} introduced attribute-based and similarity-based interfaces that enhance traditional dropdown menus with a list of font names.
Inspired by their work, we propose leveraging the FontCLIP latent space for dual-modal font retrieval tasks. 
Besides the attribute-based interface, our interface facilitates image-based retrieval, which does not require the user to obtain the vector-based font files covering all characters, unlike O'Donovan~\etal~\shortcite{o2014exploratory}.
Moreover, our interface allows for any combination of the attributes and images.
In addition, the FontCLIP latent space exhibits the capability to generalize beyond Roman fonts, allowing it to support multiple language settings.
In the following sections, we present quantitative evaluations for attribute-based, image-based, and cross-lingual font retrieval, along with qualitative evaluation of multilingual and cross-lingual font retrieval using a combination of attribute and image inputs.

\begin{table}[t!]
  \begin{subtable}{.47\linewidth}
    \centering
    \begin{tabular}{ccccc}
      \toprule
      \multicolumn{1}{c} {Model} & Accuracy $\uparrow$ \\ \midrule
      CLIP                       & 51.87\%             \\
      FontCLIP                   & 65.32\%             \\ 
      \midrule
      \rowcolor{gray!30}
      Feature-based & 65.73\% \\
      \bottomrule
    \end{tabular}
    \caption{Pairwise attribute prediction.}
    \label{tab:attribute_pred_task}
  \end{subtable}
  \begin{subtable}{.47\linewidth}
    \centering
    \begin{tabular}{ccccc}
      \toprule
      \multicolumn{1}{c} {Model} & Accuracy $\uparrow$ \\ \midrule
      CLIP                       & 67.68\%             \\
      FontCLIP               & 74.39\%             \\  
      \midrule
      \rowcolor{gray!30}
      Feature-based & 75.95\% \\
      \bottomrule
    \end{tabular}
    \caption{Pairwise similarity prediction.}
    \label{tab:similarity_pred_task}
  \end{subtable}
  \begin{subtable}{\linewidth}
    \centering
    \begin{tabular}{ccc}
      \toprule
      \multicolumn{1}{c} {Model}      & \textit{In-domain} accuracy  & \textit{Out-of-domain} accuracy     \\ \midrule
      CLIP     & $54.92\%$ & $42.26\%$ \\
      FontCLIP & $\mathbf{64.14}\%$ & $\mathbf{64.48}\%$ \\
      \midrule
      \rowcolor{gray!30}
      Feature-based & N/A & N/A \\
      \bottomrule
    \end{tabular}
    \caption{CJK fonts pairwise attribute prediction.}
    \label{tab:CJ_attr_prediction_task}
  \end{subtable}
  \caption{
    (a)(b) For both the pairwise attribute prediction task and pairwise similarity task, FontCLIP outperforms CLIP's performance and achieves similar performance to the best model that relies on geometric typographical features computed from the vector-based font file including all characters (``Feature-based'')~\cite{o2014exploratory}, while our FontCLIP-based method only requires a font image as input.
    (c) The FontCLIP latent space generalizes to \emph{out-of-domain} attributes and to multiple languages. Note that previous feature-based methods require the vector-based font files, cannot recognize \emph{out-of-domain} attributes, and might be able to support multi-lingual capabillities only if they had access to the multi-lingual fonts files (although this has never been tested).
  }
\end{table}

\subsection{Quantitative Evaluation}
To quantitatively evaluate the attribute-based and image-based font retrieval, we follow the experiment setup outlined in~\cite{o2014exploratory} and focus on \emph{in-domain} attributes.
For attribute-based retrieval, we conduct the pairwise attribute prediction task, while for image-based retrieval, we evaluate the pairwise font similarity prediction task.
In both tasks, we use the complete $200$ fonts and the $31$ adjectives attributes in the dataset from~\cite{o2014exploratory}.
Our main goal of FontCLIP is to enable font retrieval without the need to access the original vector-based font files.
Therefore, we primarily focus on comparing the performance of FontCLIP and CLIP because both methods use font images as input instead of the vector-based font files.
For reference, we provide a performance of the best machine learning model using vector-based typographical features~\cite{o2014exploratory}.

\paragraph*{Pairwise Attribute Prediction}
The goal of the pairwise attribute prediction task is to determine which font has a higher attribute score between two font options represented as font images (\cref{fig:classification_tasks}(a)).
We first evaluate this task for \emph{in-domain} attributes.%
In total, we generated $198,400$ pairwise comparison subtasks for evaluating this task.
Each comparison was assessed by seven people, and we computed the accuracy through respective comparisons.
As shown in~\cref{tab:attribute_pred_task}, 
FontCLIP achieves better performance compared to CLIP. 
This suggests that the FontCLIP feature is more distinguishable regarding different attributes.
Moreover, despite taking only a font image as input, FontCLIP achieves comparable performance to the best model that uses typographical features proposed by~\cite{o2014exploratory}.
This result suggests that FontCLIP's image-based typographical features extracted only from target glyphs (\ie~not glyphs of all Roman characters) are representative and achieve similar retrieval performance as vector-based typographical features extracted from all Roman characters.

Furthermore, we conducted quantitative evaluations to assess FontCLIP's generalization capabilities on \emph{out-of-domain} attributes and different languages.
\emph{Out-of-domain} attributes cannot be handled by the methods in~\cite{o2014exploratory} because their models need to be trained in an attribute-specific manner. 
In addition, their models use typographical features that are specifically designed for Roman characters.
For the evaluations, we additionally collected pairwise attribute rating data for $50$ CJK fonts with three participants recruited from our university.
The collected dataset contains ratings for $5$ \emph{in-domain} attributes and $3$ \emph{out-of-domain} attributes specifically used for describing CJK fonts, including ``traditional'', ``robust'', and ``Japanese style''.
In \cref{tab:CJ_attr_prediction_task}, we show the results of the attribute prediction task performed on CJK fonts.
It can be observed that FontCLIP outperforms CLIP on both \emph{in-domain} and \emph{out-of-domain} attributes in this task, especially on \emph{out-of-domain} attributes.

\paragraph*{Pairwise Similarity Prediction}
The pairwise similarity prediction task involves selecting the font that is more similar to a given reference font image out of two font images (\cref{fig:classification_tasks}(b)).
This task serves as a means to assess whether the distances in the FontCLIP latent space accurately reflect the perceptual similarity between fonts.
In total, we generated $35,387$ comparisons for this task.
Each pairwise comparison was voted by $10$ to $15$ individuals, and we computed the accuracy through respective comparisons.
For the analysis, we excluded $52$ comparisons with tie votes.
As shown in~\cref{tab:similarity_pred_task}, similar to the pairwise attribute prediction task, FontCLIP obtains better performance compared to CLIP and achieves comparable performance to the best model that uses vector-based typographical features extracted from all Roman characters~\cite{o2014exploratory}.
The results of both tasks indicate that FontCLIP's image-based typographical features perform similarly to its vector-based counterpart without requiring vector-based font files for all Roman characters, thus significantly reducing the efforts of retrieving new fonts.
\begin{figure}[t]
    \centering
    \includegraphics[width=\columnwidth]{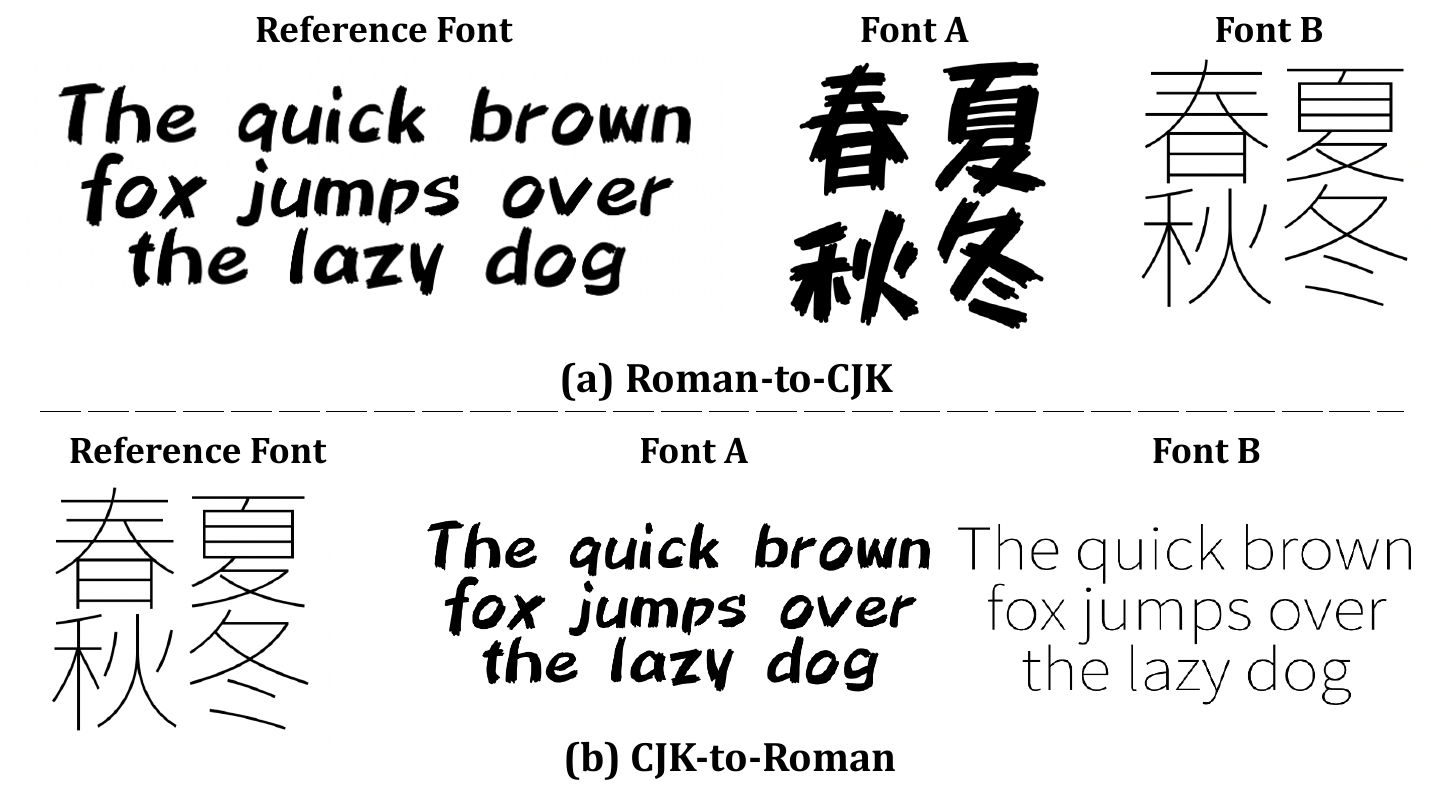}
    \caption{
    Cross-Lingual Pairwise Similarity Prediction. (a) For the ``Roman-to-CJK'' task, we use a classifier to determine which CJK font is more similar to a reference Roman font between two CJK font options.
    (b) Conversely, for the ``CJK-to-Roman'' task, we use a classifier to determine which Roman font is more similar to a reference CJK font between two Roman font options.
    In both tasks, we compare the results obtained using the classifiers with human judgments, and calculate the prediction accuracy as the performance metric.
    }
    \label{fig:crosslingual_tasks}
\end{figure}

\paragraph*{Cross-Lingual Pairwise Similarity Prediction Task}
We conduct two pairwise similarity prediction tasks to assess the cross-lingual retrieval ability of different methods, 
As shown in~\cref{fig:crosslingual_tasks}, the first task is called ``Roman-to-CJK'', which involves selecting the CJK font that is more similar to a given Roman font.
The second task is ``CJK-to-Roman'', which involves selecting the Roman font that is more similar to the query CJK font.
To evaluate these two tasks quantitatively, we collected $280$ fonts that were not part of the training dataset.
In total, we generated $100$ pairwise comparison subtasks for ``Roman-to-CJK'' and ``CJK-to-Roman'' tasks and recruited five participants to rate these comparisons.
In~\cref{tab:crosslingual}, we show the results for both ``Roman-to-CJK'' and ``CJK-to-Roman'' tasks.
We can observe that FontCLIP's predictions are better aligned with human ratings compared to CLIP's predictions.
This suggests that despite being finetuned solely on the Roman character dataset, the FontCLIP model can still learn general typographical features that achieve better cross-lingual font retrieval.
We also observed that FontCLIP's performance of ``Roman-to-CJK'' was better than its ``CJK-to-Roman'' counterpart, which is in line with our expectation given that the FontCLIP model was trained only on the Roman character dataset.
\begin{table}[t]
\centering
\begin{tabular}{cccc} 
\toprule
Model & Roman-to-CJK $\uparrow$  & CJK-to-Roman $\uparrow$  \\
\midrule
CLIP  & $57.4$\% & $50.0$\% \\ 
FontCLIP & $\mathbf{67.2}$\% & $\mathbf{62.6}$\% \\  
\bottomrule
\end{tabular}
\caption{
For both cross-lingual pairwise similarity tasks, FontCLIP performed better than CLIP, suggesting that FontCLIP's prediction results are closer to human rating results.
}
\label{tab:crosslingual}
\end{table}

\begin{figure*}[h]
\centering
  \includegraphics[width=\textwidth]{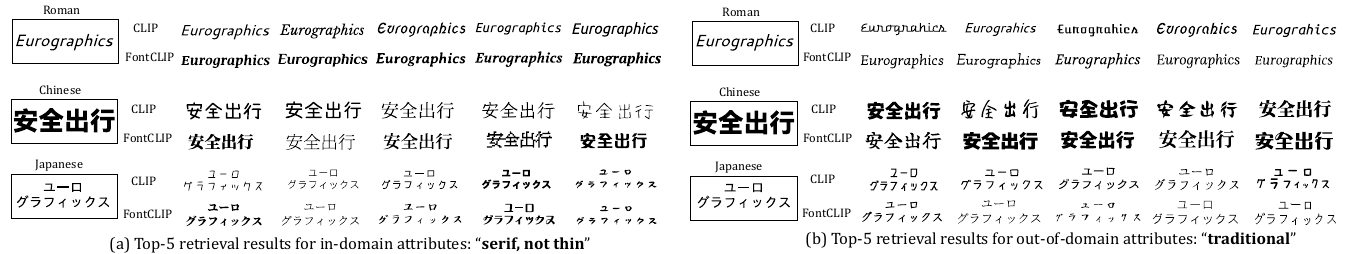}
  \caption{
  The results of dual-modal font retrieval using FontCLIP latent space and CLIP latent space.
  The goal of this multi-modal retrieval is to preserve the style of input font image query (text with frame) while incorporating the desired attributes.
  (a) We show the top-5 retrieval results with \emph{in-domain} attributes for Roman, Chinese, and Japanese characters.
  By using the FontCLIP feature, we can retrieve more fonts with serif for multiple languages.
  (b) We show the top-5 retrieval results with \emph{out-of-domain} attributes for Roman, Chinese, and Japanese characters.
  }
\label{fig:retrieval_result}
\end{figure*}
\begin{figure*}[h]
\centering
  \includegraphics[width=\textwidth]{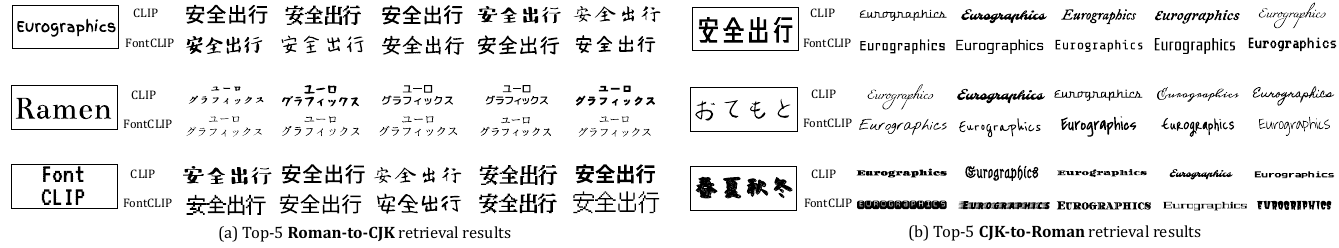}
  \caption{
  Cross-lingual font retrieval results using FontCLIP latent space and CLIP latent space.
  The goal of this cross-lingual font retrieval is to find fonts in other languages that match the style of the input font image query (text with frame). 
(a) We show the top-5 retrieval results for ``Roman-to-CJK''. (b) We show the top-5 retrieval results for ``CJK-to-Roman''.
  }
\label{fig:crosslingual_qual}
\end{figure*}

\subsection{Qualitative Evaluation}
In our qualitative evaluation, we collected $1,169$ Roman fonts and $293$ CJK fonts in total.
For each font, we generate its font image using the method described in~\mbox{\cref{sec:font_image}} and extract its visual feature using FontCLIP visual encoder $E_{I}$.
We denote the final font feature databases for Roman and for CJK as $\Omega_{\text{Roman}}$ and $\Omega_{\text{CJK}}$.

\paragraph*{Multilingual Font Retrieval}
First, we demonstrate FontCLIP's unprecedented generalization capability by showing multilingual font retrieval results using a combination of attributes and image inputs.
Our goal aligns with~\cite{Tugba2020Happy}, where we aim to retrieve results that preserve the style of input font image while incorporating the desired attributes.
Specifically, given a query font image $I_{\text{query}}$ and a set of desired attributes $\mathbf{A}=\{a_1, a_2,..., a_N\}$, we obtain the embedding vector of the desired font $e_{\text{desired}}$ using the following formulation:
\begin{align}
e_{\text{desired}} = E_{I}(I_{\text{query}})+ w E_{T} (T),
\end{align}
where $E_{I}$ and $E_{T}$ is the image and text encoder of FontCLIP, $T$ is a text prompt containing all desired attribute $\mathbf{A}$, and $w\in [0,1]$ is a weight that controls the balance between preserving the original letter styles and incorporating the styles of the desired attributes.
With the embedding vector $e_{\text{desired}}$, we obtain the top-$K$ retrieved results by choosing the $K$ closest fonts to $e_{\text{desired}}$ in the corresponding font feature database ($\Omega_{\text{Roman}}$ or $\Omega_{\text{CJK}}$). 

In~\cref{fig:retrieval_result}, we compared the retrieved results obtained using FontCLIP and CLIP in different languages, considering both \emph{in-domain} and \emph{out-of-domain} attributes.
In~\cref{fig:retrieval_result}, we found that the retrieved results using FontCLIP effectively incorporate the desired attributes while preserving the original style. 
For example, in~\cref{fig:retrieval_result}(a), more retrieved results are with serifs for all languages by using the FontCLIP feature than that of CLIP;
in~\cref{fig:retrieval_result}(b), for \emph{out-of-domain} attribute ``traditional'', the retrieved results by FontClip also better align with the human perception than that of CLIP.
Besides, the CLIP latent space fails to interpret the ``not'' prompt, resulting in thicker retrieved results compared to the results obtained by the FontCLIP feature (\cref{fig:retrieval_result}(a)).

\paragraph*{Cross-Lingual Font Retrieval}
Next, we demonstrate the cross-lingual font retrieval results using FontCLIP.
Our goal is to retrieve fonts that have a similar style to the query font image $I_{\text{query}}$ from other languages.
To begin with, we calculate the visual feature of the query font image $I_{\text{query}}$ as $E_I(I_{\text{query}})$.
Following this, we search for the $k$ nearest fonts to the visual feature in the font feature dataset of other languages.
In the ``Roman-to-CJK'' results shown in~\cref{fig:crosslingual_qual}, we have found that the FontCLIP feature is better at retrieving CJK fonts that are more similar to the query Roman fonts, compared to the CLIP feature (\cref{fig:crosslingual_qual}(a)).
Meanwhile, we observed that the ``CJK-to-Roman'' retrieved results of CLIP deviate excessively from the style of the query font image (\cref{fig:crosslingual_qual}(b)).%
\begin{figure}[t!]
    \centering
    \includegraphics[width=\linewidth]{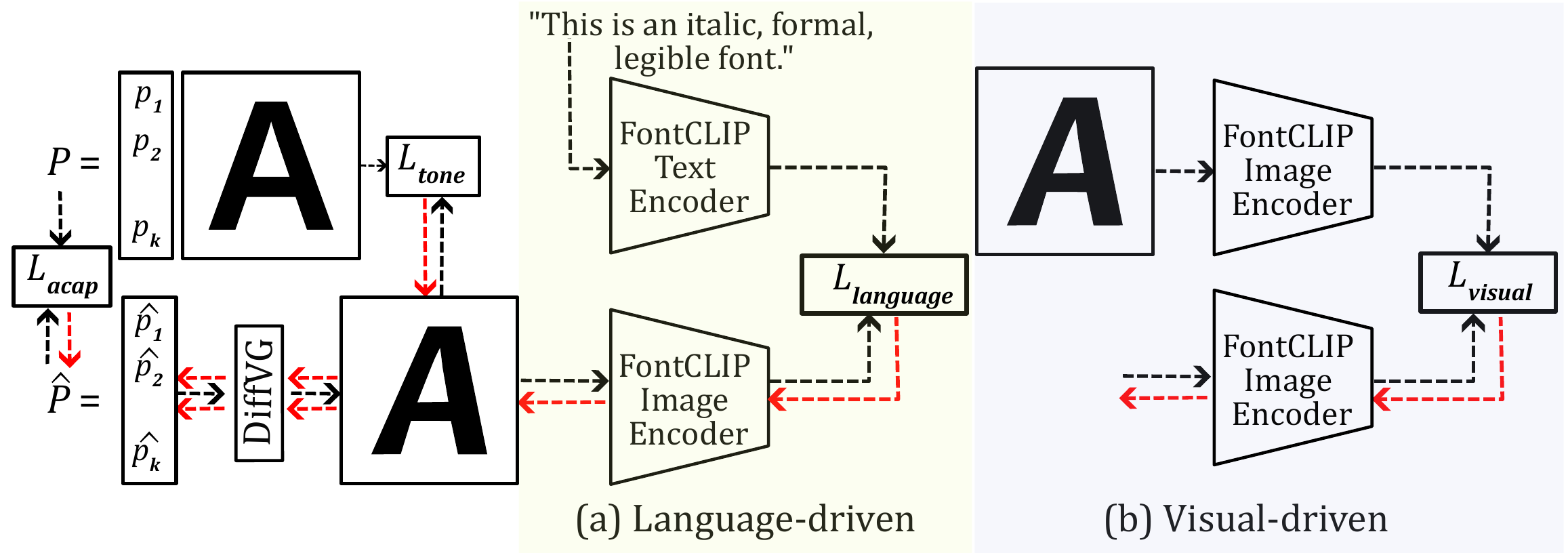}
    \caption{
    An overview of our multi-modal vector font optimization. 
    Given an input letter $l$ (``A'' in this example)  represented as a set of outline control points $P$, and either a language-driven descriptive prompt $T_{\text{user}}$ (a), or a visual-driven reference font image $I_{\text{user}}$ (b), we iteratively optimize the new positions of $\hat{P}$ creating the optimized letter shape $\hat{l}$. 
    Inspired by~\cite{iluz2023wordasimage}, we first rasterize the deformed letter $\hat{l}$ by a differentiable rasterizer (DiffVG).
    To guide the optimization, we use a language loss $L_{\text{language}}$ in (a) language-driven optimization, or a visual loss $L_{\text{visual}}$ in (b) visual-driven optimization to ensure  $\hat{l}$ aligns with desired attributes indicated by the descriptive prompt or the reference font image.
    Moreover, our objective function includes the tone preservation loss $L_{\text{tone}}$ and an ACAP deformation loss $L_{\text{acap}}$ similar to~\cite{iluz2023wordasimage}.
    Black and red dashed arrows indicate forward and backward computation, respectively.
    }
    \label{fig:opt_framework}
\end{figure}

\begin{figure*}[!h]
    \centering\includegraphics[width=\linewidth]{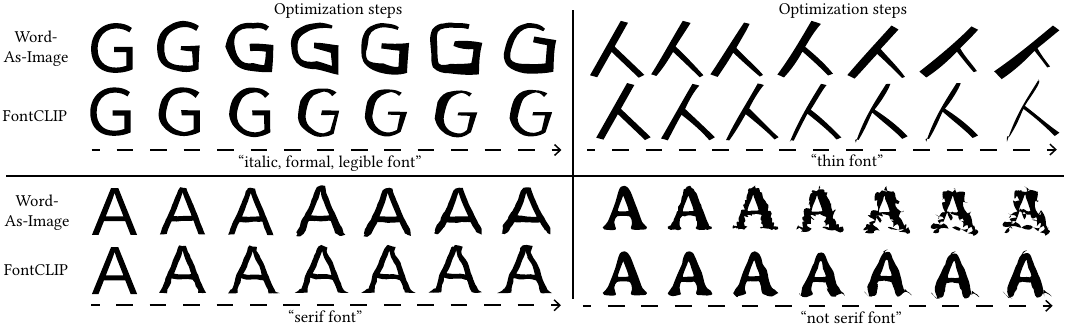}
    \caption{
    Visualization of the vector font optimization steps of the language-driven Roman and Chinese character optimization using FontCLIP. \minorhl{We compared the results obtained by Word-As-Image~\mbox{\cite{iluz2023wordasimage}} and our method.    
    Our method better captures and reconstructs each character's typographical features, including features such as serif.}
    }
\label{fig:opt_text_driven}
\end{figure*}

\begin{figure}[!h]
\centering
  \includegraphics[width=\columnwidth]{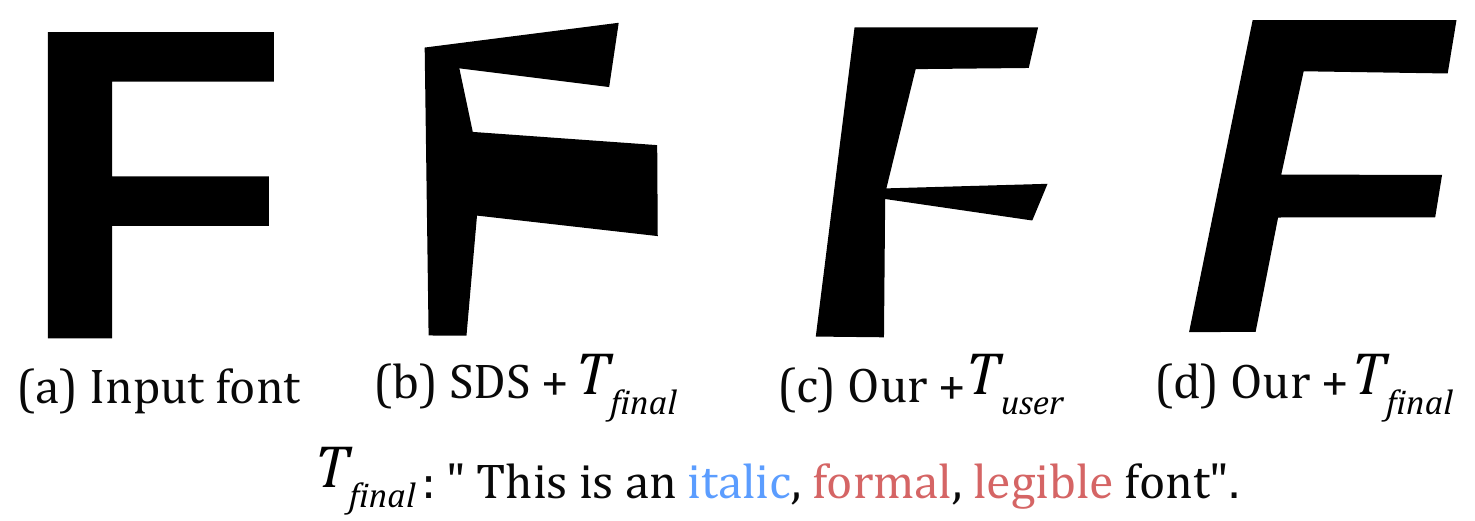}
  \caption{
  Ablation study on the language-driven font optimization.
  Given (a) an input font, we compare the results obtained by (b) replacing $L_{\text{language}}$ into SDS loss, (c) our method using only $T_{\text{user}}$, and (d) our method using $T_{\text{final}}$.
  (The user specificed attributes are shown in \textcolor{softblue}{blue} and the attributes to be preserved are shown in \textcolor{softred}{red}.)
  }
\label{fig:opt_ablation}
\end{figure}

\section{Dual-Modal Multilingual Vector Font Optimization}

In this section, we describe another application of FontCLIP: an optimization-based method that modifies the letter shapes in vector fonts based either on text prompts or on image inputs in multiple languages.
\cref{fig:opt_framework} shows the overview of our vector font optimization method.
Guided by the FontCLIP latent space, our optimization-based method supports both language-driven and image-driven font optimization in multiple languages.

The input to our method is a letter $l$ from an existing font in vector format.
Following~\cite{iluz2023wordasimage}, we represent $l$ as a set of $k$ control points, $P = \{ p_{j} \in \mathbb{R}^{2} \}_{j = 1}^{k}$, describing it's outline. 
$P$ is obtained by a subdivision, which provides sufficient expressiveness even for letters with few control points.
The output of our pipeline is the same set of control points $\hat{P} = \{ \hat{p}_{j} \}_{j = 1}^{k}$ in different positions that represents the outline of the manipulated letter $\hat{l}$.
The users define their goal either by providing a text prompt $T_{\text{user}}$ of attributes or a reference font image $I_{\text{user}}$ as additional input that drives the optimization (see \cref{fig:opt_framework}).

\subsection{Language-Driven Font Optimization}
Our specific goal here is to manipulate the original letter $l$ into $\hat{I}$ by aligning it with desired attributes while preserving the original styles.
To preserve the original styles of $l$, we begin by calculating similarity scores between its original shape $\mathcal{R}(P)$ and the $37$ attributes.
We then select the top-$M$ (we set $M=2$) attributes with the highest similarity scores as the attributes to be preserved.
The user-specified prompt $T_{\text{user}}$ is then combined with these $M$ preserved attributes to form the final compound descriptive prompt $T_{\text{final}}$. 
To encourage the manipulated letter $\hat{l}$ to align with $T_{\text{final}}$, we define the following function using the FontCLIP visual encoder $E_I$ and text encoder $E_T$:
\begin{align}
L_{\text{language}}(\hat{P}, T_{\text{final}}) = \text{dist}(E_I(\mathcal{R}(\hat{P})), E_T(T_{\text{final}})),
\end{align}
where $\text{dist}(\mathbf{x},\mathbf{y})=1.0-\frac{\mathbf{x}\cdot\mathbf{y}}{\|\mathbf{x}\|\|\mathbf{y}\|}$ denotes the cosine distance between $\mathbf{x}$ and $\mathbf{y}$, and $\mathcal{R}$ is the differentiable rasterizer~\cite{Li2020DVG}.
However, we have noticed that using $L_{\text{language}}$ alone can result in significant deviations from the initial letter geometry.
Inspired by~\cite{iluz2023wordasimage}, we incorporate the ACAP deformation loss and tone preservation loss into our final objective function.
The ACAP deformation loss minimizes the deviation of the final letter shape from its initial shape, while the tone preservation loss aims to preserve the font's style and letter structure.
For detailed definitions of both losses, please refer to~\cite{iluz2023wordasimage}.
The objective function is defined as
\begin{align}
L_{LD} = L_{\text{language}} + w_{\text{acap}}L_{\text{acap}} +  w_{\text{tone}}L_{\text{tone}},
\end{align}
where we set $w_{\text{acap}}=0.2$ and $w_{\text{tone}}=0.2$ throughout all examples shown in this paper.
In~\cref{fig:teaser}(b), we can observe the iterative optimization steps where the Chinese character gradually becomes thinner while maintaining its formal and legible appearance.
In~\cref{fig:opt_text_driven}, we \minorhl{compare the results obtained by our method and Word-As-Image\mbox{~\cite{iluz2023wordasimage}} on Roman and Chinese characters.}
\minorhl{We can observe that our optimization method using FontCLIP feature captures and reconstructs typographical features better, even for features such as serif.
}

\paragraph*{Ablation Study}
In~\cref{fig:opt_ablation}, we present a comparison of various formulations.
This includes replacing the language loss $L_{\text{language}}$ with the Stable Diffusion (SDS) loss, which was used in~\cite{iluz2023wordasimage}, and solely using $T_{\text{user}}$ in $L_{\text{language}}$ (\ie~excluding the attributes we aim to preserve). 
As can be seen, the result using SDS loss did not exhibit the desired attributes used in the text prompt $T_{\text{final}}$ and severely deviated from the original letter shape.
The result using $T_{\text{user}}$ reflects the desired attribute (``italic'') but fails to preserve the original styles of the input font.

\subsection{Image-Driven Font Optimization}
\begin{figure}[!t]
        \centering\includegraphics[width=\linewidth]{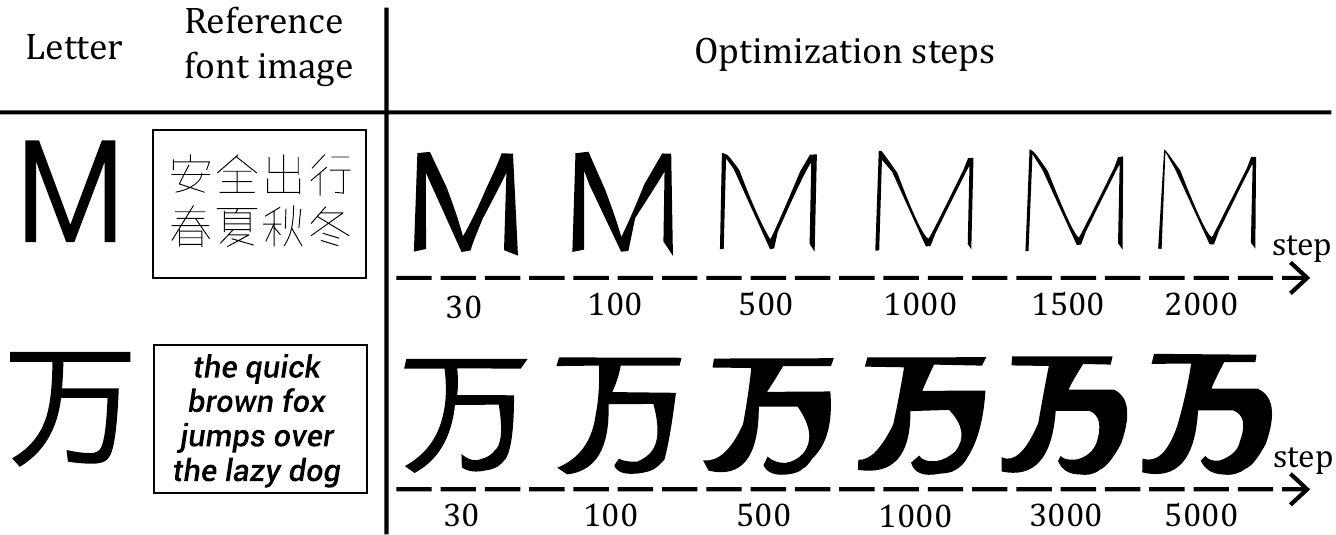}
        \caption{
        Visualization of the optimization steps of the cross-lingual image-driven Roman and Chinese character optimization.}
        \label{fig:opt_image_driven}
\end{figure}

\begin{figure}[!htb]
        \centering
        \includegraphics[width=\linewidth]{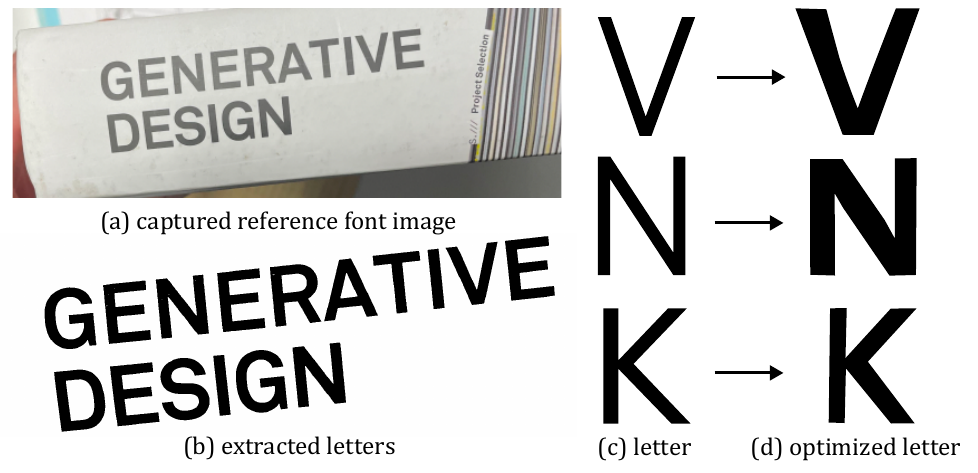}
        \caption{(a) Given a reference font image captured in real-world, our optimization method uses (b) the extracted letters to  manipulate (c) the input letters.
        (d) The optimized letters exhibits a similar style to the fonts in the captured image.}
        \label{fig:opt_billboard}
\end{figure}

When a reference font image $I_{\text{user}}$ is given, image-driven font optimization manipulates $l$ into $\hat{l}$ while ensuring that $\hat{l}$ reflects the visual typographic attributes present in $I_{\text{user}}$.
To achieve this, we define the following function using the FontCLIP visual encoder $E_I$ and text encoder $E_T$:
\begin{align}
L_{\text{image}}(\hat{P}, I_{\text{user}}) = \text{dist}(E_I(\mathcal{R}(\hat{P})), E_I(I_{\text{user}})),
\end{align}
and the overall objective function for image-driven font manipulation is defined as:
\begin{align}
L_{VD} = L_{\text{image}} + w_{\text{acap}}L_{\text{acap}} +  w_{\text{tone}}L_{\text{tone}},
\end{align}
where we set $w_{\text{acap}}=0.2$ and $w_{\text{tone}}=0.2$.
In~\cref{fig:opt_image_driven}, we present cross-lingual optimization results on both Roman and CJK characters.
We can observe the iteratiove optimization steps that gradually align the styles of the input Roman and Japanese letters with the style in the reference font images even from other languages.
Finally, in~\cref{fig:opt_billboard}, we demonstrate the effectiveness of our font optimization method using a reference font image captured in real-world conditions.
We extracted the letters from the captured image and used them as $I_{\text{user}}$ to drive the optimization for the provided letters.
\begin{figure}[h]
    \centering
    \includegraphics[width=\linewidth]{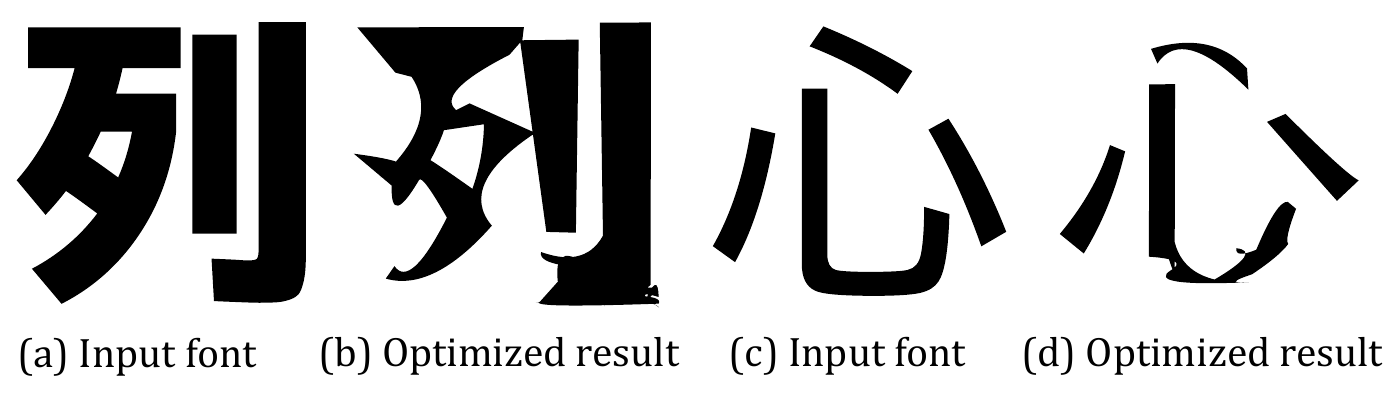}
    \caption{
Our optimization method faces challenges in handling complex structures such as crossing and rounded strokes.
    }
    \label{fig:cj_failure}
\end{figure}
\section{Limitations and Future Work}
\paragraph*{Attribute Entanglement}
Currently, FontCLIP latent space exhibits entanglement between different attributes.
As shown in~\cref{fig:opt_ablation}(c), the optimized letter exhibits characteristics from attributes that are not specificed by the user.
As a result, our method need to identify and preserve the most representative attributes of the font during language-driven font optimization (\cref{fig:opt_ablation}(d)).
In the future, a potential solution would be to explore contrastive finetuning, utilizing fonts with similar attribute scores but differing in only one attribute.

\paragraph*{Vector Font Optimization on Complex Typographic Structures}
While our current character shape optimization method shows promising results, it faces challenges in handling complex typographic structures, such as crossing and rounded strokes (\cref{fig:cj_failure}).
Future research could improve our optimization method by investigating more appropriate font parameterizations~\cite{Tamir2010PFont} and incorporating more typographic-specific constraints.
Nonetheless, our results validate the concept and suggest that our FontCLIP could be a foundation of future font optimization methods.

\paragraph*{Generalization Enhancement}
Currently, FontCLIP is specifically finetuned using a dataset that exclusively contains Roman alphabet characters and commonly associated attributes.
\minorhl{However, there is a possibility that cultural differences might affect how letter shapes are linked to attributes.}
\minorhl{To alleviate this issue}, we plan to explore few-shot learning techniques for \emph{out-of-domain} languages, which involve collecting small-scale datasets using the data collection process described in~\cite{o2014exploratory}.
\section{Conclusion}
In this paper, we introduced FontCLIP – a model that bridges the semantic understanding of a large vision-language model with typographical knowledge. 
Our experiments demonstrated FontCLIP's two unprecedented generalization abilities.
First, FontCLIP can generalize to multiple languages despite being finetuned only on a Roman character dataset.
This ability enables multilingual and cross-lingual font retrieval and letter shape optimization.
Second, FontCLIP can recognize \emph{out-of-domain} semantic attributes, facilitating more diverse attribute-based font retrieval and letter shape optimization.
Finally, FontCLIP's dual-modality allows unprecedented multilingual font applications through a unified space without extracting typographical features through vector-based font files.
In summary, we believe FontCLIP can greatly simplify the process of obtaining desired fonts during the design process.
\section*{Acknowledgement}
We thank the anonymous reviewers for their valuable feedback.
This work was partially supported by JST AdCORP, Grant Number JPMJKB2302, JSPS Grant-in-Aid JP23K16921, Japan, and a collaboration with Dentsu Digital.


\bibliographystyle{eg-alpha-doi} 
\bibliography{paper}       

\end{document}